\documentclass{article}

\makeatletter
\newcommand\tabcaption{\def\@captype{table}\caption}
\newcommand\figcaption{\def\@captype{figure}\caption}
\makeatother

\usepackage{microtype}
\usepackage{graphicx}
\usepackage{subfigure}
\usepackage{booktabs} 

\usepackage{xcolor, colortbl}
\definecolor{mygray}{gray}{0.92}

\usepackage{hyperref}

\usepackage[preprint]{icml2023}

\usepackage{amsmath}
\usepackage{amssymb}
\usepackage{mathtools}
\usepackage{amsthm}

\usepackage{subcaption}
\usepackage{float}
\usepackage{bbm}
\usepackage{array}
\usepackage{pifont}
\usepackage{setspace}
\usepackage{makecell}
\usepackage{multirow}
\usepackage{multicol, bbding}
\usepackage[capitalize,noabbrev]{cleveref}

\theoremstyle{plain}

\theoremstyle{definition}

\theoremstyle{remark}

\usepackage[textsize=tiny]{todonotes}

\begin{document}

\twocolumn[
\icmltitle{DomainVerse: A Benchmark Towards Real-World Distribution Shifts For Tuning-Free Adaptive Domain Generalization}



\icmlsetsymbol{equal}{*}

\begin{icmlauthorlist}
\icmlauthor{Feng Hou}{equal,ict,ucas}
\icmlauthor{Jin Yuan}{equal,seu}
\icmlauthor{Ying Yang}{research}
\icmlauthor{Yang Liu}{ict,ucas}
\icmlauthor{Yang Zhang}{research}
\icmlauthor{Cheng Zhong}{research}
\icmlauthor{Zhongchao Shi}{research}
\icmlauthor{Jianping Fan}{research}
\icmlauthor{Yong Rui}{research}
\icmlauthor{Zhiqiang He}{ict,ucas,lenovo}
\end{icmlauthorlist}

\icmlaffiliation{ict}{Institute of Computing Technology (ICT), Chinese Academy of Sciences}
\icmlaffiliation{ucas}{University of Chinese Academy of Sciences}
\icmlaffiliation{seu}{Southeast University}
\icmlaffiliation{research}{Lenovo Research}
\icmlaffiliation{lenovo}{Lenovo Ltd.}

\icmlcorrespondingauthor{Feng Hou}{houfeng19@mails.ucas.ac.cn}

\icmlkeywords{Adaptive Domain Generalization, DomainVerse}

\vskip 0.3in
]



\printAffiliationsAndNotice{\icmlEqualContribution} 

\begin{abstract}
Traditional cross-domain tasks, including domain adaptation and domain generalization, rely heavily on training model by source domain data. With the recent advance of vision-language models (VLMs), viewed as natural source models, the cross-domain task changes to directly adapt the pre-trained source model to arbitrary target domains equipped with prior domain knowledge, and we name this task Adaptive Domain Generalization (ADG). However, current cross-domain datasets have many limitations, such as unrealistic domains, unclear domain definitions, and the inability to fine-grained domain decomposition, which drives us to establish a novel dataset DomainVerse for ADG. Benefiting from the introduced hierarchical definition of domain shifts, DomainVerse consists of about 0.5 million images from 390 fine-grained realistic domains. With the help of the constructed DomainVerse and VLMs, we propose two methods called Domain CLIP and Domain++ CLIP for tuning-free adaptive domain generalization. Extensive and comprehensive experiments demonstrate the significance of the dataset and the effectiveness of the proposed methods.
\end{abstract}

\section{Introduction}
\label{intro}

Deep neural networks usually rely on the assumption that training data and test data are independent and identically distributed~\cite{zhou2021domainsuvery}, which greatly hinders their further development into realistic scenarios~\cite{zhou2023mixstyle}. To tackle the aforementioned challenge, domain generalization (DG)~\cite{QinweiXu2021AFF, hou2023learning} is proposed to train a source model on multi-source domains that can generalize to arbitrary target domains. Unlike DG, unsupervised domain adaptation (UDA)~\cite{zhou2021domain, yuan2022self} is introduced to train a specific model based on single or multiple source domains with unlabelled target domain data. 
\begin{figure}[t]
    \centering
    \includegraphics[width=\linewidth]{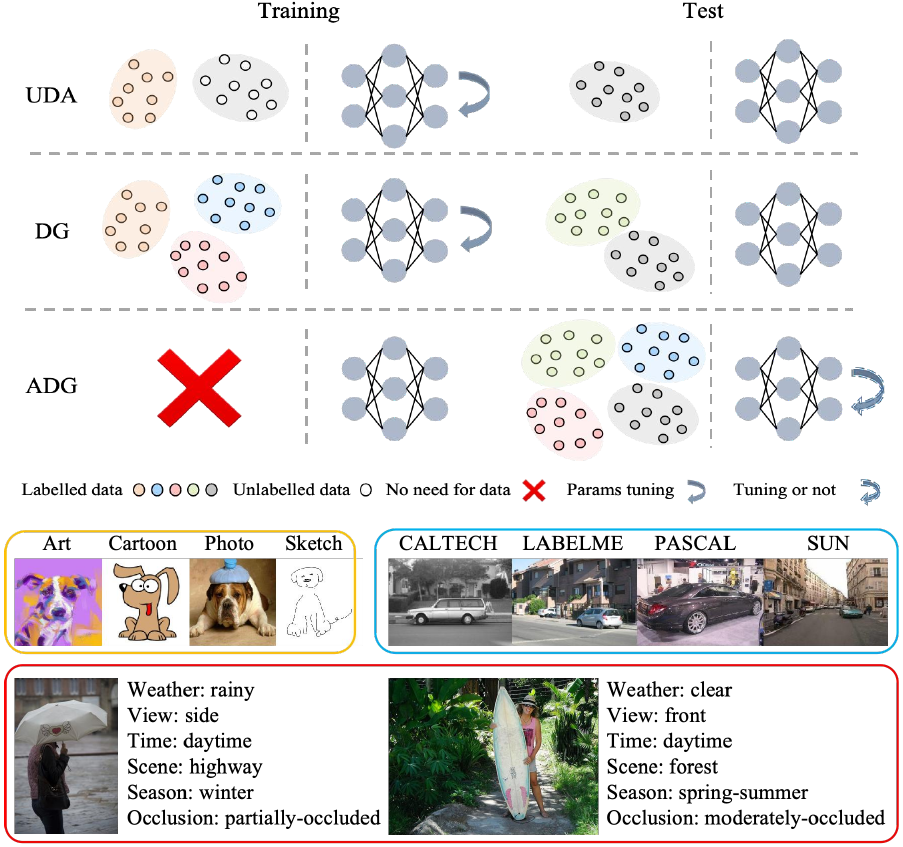}
	\caption{In the upper part, we demonstrate the paradigm of the ADG task and traditional UDA and DG. In the middle part, we present the issues of unrealistic domains and unclear domain definitions in the current cross-domain datasets. In the bottom part, we display the common real-world domain shifts to illustrate the difficulty of decoupling these characteristics. }
	\label{fig:dgdata}
\end{figure}
\begin{table*}[t]
\centering
\resizebox{0.95\linewidth}{!}{
\begin{tabular}{c|ccc|cccc}
\toprule
Dataset & Domain shifts (Definition)  & Domains & Images & Realistic & Hierarchical & Decomposable & Balanced   \\ \midrule
PACS & 1 (Style) & 4 & 9,991 & \XSolidBrush & \XSolidBrush & \XSolidBrush & \XSolidBrush \\
Office-Home & 1 (Style) & 4 & 15,913 & \XSolidBrush & \XSolidBrush & \XSolidBrush & \XSolidBrush \\
Digits-Five & 1 (Color) & 5 & 145,298 & \XSolidBrush & \XSolidBrush & \XSolidBrush & \XSolidBrush \\
DomainNet & 1 (Style) & 6  & 569,010 & \XSolidBrush & \XSolidBrush & \XSolidBrush & \XSolidBrush  \\
VLCS & N/A & 4 & 10,729 & \Checkmark & \XSolidBrush & \XSolidBrush & \XSolidBrush \\
OOD-CV & 5 (Shape, Pose, Texture, Context, Weather) & 5 & 13,297 & \Checkmark & \XSolidBrush & \XSolidBrush & \XSolidBrush \\
NICO$^{++}$ & 3 (Location, Background, Time) & 10 & 232,400 & \Checkmark & \Checkmark & \XSolidBrush & \XSolidBrush \\
DomainVerse (Ours) & 5 (Weathers, Views, Time, Seasons, Occlusion) & 390 & 457,570 & \Checkmark & \Checkmark & \Checkmark & \Checkmark \\
\bottomrule
\end{tabular}}
\caption{Comparison of our proposed DomainVerse and existing cross-domain datasets.}
\label{tab:compar}
\end{table*}
Recent advancements in vision-language foundation models (VLMs), such as CLIP~\cite{radford2021learning} and ALIGN~\cite{jia2021scaling}, recognized as a well-generalized pre-trained model, have paved the way for natural source model. In this context, we propose a groundbreaking paradigm named Adaptive Domain Generalization (ADG). Illustrated in Figure~\ref{fig:dgdata}, ADG strategically combines the strengths of both DA and DG that not only exploit specific domain knowledge but also generalize well to arbitrary unseen domains. In summary, the novel task is designed to evaluate the VLM's ability across arbitrary target domains only leveraging the prior knowledge from the target domain during inference.

To adequately exploit the strong power of VLMs in ADG task, a benchmark with sufficient diversity of domains for real-world scenarios is urgent. However, the mainstream domain datasets suffer from following limitations: 1. Prevailing datasets often lack realism, making adaptation to real-world scenarios challenging (e.g., color for Digits-five \cite{hull1994database, lecun1998gradient, netzer2011reading, ganin2016domain}, and style for PACS~\cite{li2017deeper}, Office-Home~\cite{venkateswara2017deep}, and DomainNet~\cite{peng2019moment}). 2. Some datasets suffer from an inadequate number of domains (e.g., 4 domains in  Office-Home~\cite{venkateswara2017deep} and 6 in DomainNet~\cite{peng2019moment}). 3. Current datasets often lack clear domain definitions and fail to provide proper measures for distribution shifts (e.g., VLCS~\cite{torralba2011unbiased}).  4. High cost of data collection and annotation results in insufficient sample size, incorrect labelling, and imbalanced data problems (e.g., NICO$^{++}$~\cite{zhang2023nico++}, and OOD-CV~\cite{zhao2022ood}). In real-world scenarios, a single image often undergoes multiple domain combinations from different domain shifts, as depicted in Figure~\ref{fig:dgdata}. This observation motivates us to establish a clear decomposable definition of realistic domain shifts. However, collecting a large-scale realistic dataset is challenging due to the high cost of decoupling domain shift combinations. 

To overcome these challenges, we leverage the highly controllable tools available in a game-engine-based simulator\footnote{\url{https://unity.com}}~\cite{haas2014history,hussain2020unity} to synthesize images that faithfully simulate realistic scenarios. As a result, we construct a novel large-scale cross-domain dataset named DomainVerse ($\sim 0.5$ million images). This dataset comprises 5 domain shifts frequently observed in real-world situations, with each shift consisting of corresponding distinct domains, totalling 18 coarse domains and 390 fine-grained domain combinations. Benefiting from the precision of the game engine, the synthetic images across different domains are balanced and correctly labelled. The comparison of DomainVerse and existing datasets is shown in Table~\ref{tab:compar}.

The primary challenge in applying VLMs to downstream tasks arises from the discrepancy between the distribution of the pre-trained model and the downstream dataset. Recent efforts tackle this issue through prompt tuning~\cite{zhou2022learning, zhou2022conditional}. While such methods can discover better prompts than the original ones (e.g., ``a photo of a \{class\}"), the high cost associated with fine-tuning and the requirement for labelled training data poses significant obstacles to further development in real-world applications. Alternatively, some researchers have explored zero-shot learning methods, such as CALIP~\cite{guo2023calip} and CLIP-DN~\cite{zhou2023distribution}, which enhance performance through post-processing of encoded embeddings without any parameter tuning. Besides, the introduction of prior knowledge from downstream tasks is proven as an effective way to eliminate the distribution shifts~\cite{muralidhar2018incorporating, zhou2021domain}. With the distinct domain information of DomainVerse, serving as prior knowledge of zero-shot learning, ADG can be achieved to enhance the VLMs' performance on target tasks. Therefore, we propose two tuning-free ADG methods by leveraging domain information, named Domain CLIP and Domain++ CLIP. Specifically, Domain CLIP inserts the domain name into the original prompt as the prior domain information, facilitating the model in mitigating noise stemming from diverse domains and focusing on object-centric analysis. To further utilize the rich context provided by additional linguistic information, Domain++ CLIP leverages detailed descriptions of domains to offer distinguishing features across domain shifts.

We conduct comprehensive experiments on the proposed DomainVerse to evaluate and benchmark recent vision-language models. The proposed ADG baseline Domain CLIP and Domain++ CLIP achieve a significant improvement compared with other competitive works.

\section{Related Work}
\label{works}

\noindent \textbf{Cross-Domain Datasets.}
The mainstream cross-domain datasets include Digits-five \cite{hull1994database, lecun1998gradient, netzer2011reading, ganin2016domain}, Office-Caltech10 \cite{gong2012geodesic}, Office-31 \cite{saenko2010adapting}, Office-Home \cite{venkateswara2017deep}, PACS \cite{li2017deeper}, VLCS~\cite{torralba2011unbiased}, DomainNet series~\cite{peng2019moment, zhou2021domain}, NICO$^{++}$~\cite{zhang2023nico++}, and OOD-CV~\cite{zhao2022ood}.

Digits-five is tailored for handwritten digit recognition, specifically addressing domain shifts related to color variation. Moreover, both Office-Home and PACS consist of four domains, yet their applicability to real-world scenarios is limited. For instance, the \emph{Art Painting} of PACS and \emph{Clipart} in Office-Home are distributed from \emph{style} domain shift, which fails to fully capture the distribution shifts encountered in real-world settings. The DomainNet series comprises a vast collection of images, containing domains such as \emph{Clipart}, \emph{Infograph}, \emph{Painting}, \emph{Quickdraw}, \emph{Real}, and \emph{Sketch}. While sharing the same issues with Office-Home and PACS, these datasets also suffer from limited domains compared with the realistic distribution variances. 

Other mainstream datasets, including Office-Caltech10, Office-31, and VLCS, collect images from the real world. However, their domain differences primarily arise from variations in the cameras used to capture images from different scenes. Although these datasets simulate realistic domain shifts, their domain definitions are not clear; for example, \emph{Amazon} in Office-31 and \emph{PASCAL} in VLCS can not reflect the accurate domain information. NICO$^{++}$, an extensive cross-domain benchmark collected from real-world photos, still faces imbalanced issues across domains due to scarcity in specific domains and classes in real scenarios.

Based on these observations, we construct the DomainVerse benchmark for cross-domain research, which includes decomposable realistic domain shifts, providing a more comprehensive evaluation of model performance under various domain shifts commonly encountered in the real world.

\noindent \textbf{Vision-Language (VL) models.}
In recent years, vision-language models have made a great breakthrough for zero-shot generalization~\cite{du2022survey, chen2023vlp} benefited from  CLIP~\cite{radford2021learning}, which is pre-trained on 400M image-text pairs collected from the web via contrastive learning to align the query images and the embedded words. Based on a similar setting, other works~\cite{jia2021scaling, singh2022flava} subsequently achieve further improvement in joint image-text representation learning with the help of large-scale corpus. Compared with traditional methods, the advanced mechanism facilitates the pre-trained model to realize open-vocabulary recognition. The following works also achieve strong performance in other downstream tasks, such as open-vocabulary detection~\cite{zhong2022regionclip} and segmentation~\cite{kirillov2023segment}.

\begin{figure*}[t]
    \centering
    \includegraphics[width=0.98\linewidth]{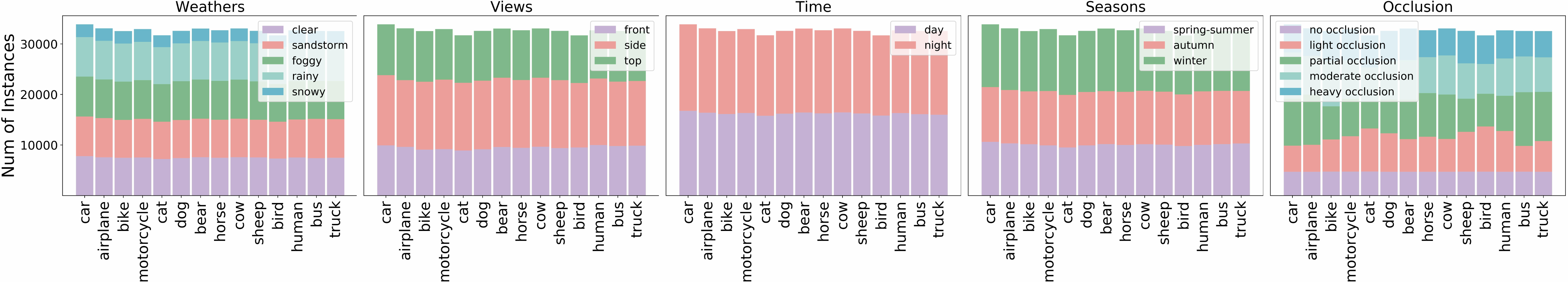}
	\caption{The DomainVerse statistics of different shifts and domains are shown in this figure. Notably, most domains achieve 
    a balanced distribution, except for \emph{snowy} solely exists in winter, and occlusion achieves relative balance due to different object sizes. The detailed statistics are shown in the Appendix.}
	\label{fig:statistic}
\end{figure*}

\noindent \textbf{Adaptation of VL models.}
While CLIP~\cite{radford2021learning} achieves promising performance on zero-shot recognition, concurrent works explore numerous mechanisms further to develop the latent potential of CLIP for downstream tasks. CALIP~\cite{guo2023calip} introduces a parameter-free module to bridge the image and text modalities during inference. Test-time distribution normalization is introduced in CLIP-DN~\cite{zhou2023distribution} to tackle the misalignment of the criterion between pre-training and zero-shot recognition. The above methods concentrate on the post-processing of encoded image and text embeddings, neglecting the impact of utilising prior knowledge for downstream tasks. VisDesc~\cite{menon2022visual} and CuPL~\cite{pratt2023does} replace the vanilla category name with the discriminative features, generated by the pre-trained Large Language Models (LLMs). These methods introduce additional context information to help encoders concentrate on objects for classification. However, the adapted models are hard to transfer to downstream tasks whose domain distributions diverge significantly from the ones during pre-training. To tackle this problem, TPT~\cite{shu2022test} is proposed to adapt the pre-trained model to arbitrary test samples on the fly by tuning the prompt and achieving moderate results. However, the increased cost of test-time prompt tuning is non-negligible. Our algorithm, tuning-free adaptive domain generalization, assimilates the advantages of recent works that adapt to arbitrary test samples without any fine-tuning only by introducing downstream domain knowledge.

\section{DomainVerse}
\label{dataset}
In this section, we introduce the design of the novel large-scale benchmark towards real-world distribution shifts, DomainVerse, and discuss the data collection and annotation.

\subsection{Hierarchical and Decomposable Domain Shifts}
For cross-domain datasets, there exists a lack of explicit definitions for both domain shifts and domains. Either these datasets fail to specify the domain shifts encompassed or incorporate a mixture of disparate domains, making decoupling unattainable. This ambiguity hinders the ability to evaluate which domain shifts a well-trained model excels at, rendering the algorithm challenging to apply in real-world scenarios. We first strictly define a variety of common domain shifts in the real world, and then several exclusive domains of each domain shift are discreetly selected. Besides, we further leverage the game engine to achieve arbitrary combinations and decoupling of different domains to simulate realistic scenarios in which one image has multiple domains from different domain shifts. Benefiting from the hierarchical structure and the decomposable attribute, the proposed DomainVerse is suitable for evaluating the adaptability of models to diverse realistic domains.

\subsection{Real-world Domains}
To construct an unbiased benchmark, we select the domains and domain shifts from outdoors, while the indoors highly depending on the scenes often introduce additional bias. Under prevalent distribution shifts observed in the real world, we designed two primary groups of domain shifts.\footnote{Note that the attributes of objects are set as the variance of DomainVerse, not a specific domain shift.}.

The first group predominantly involves environmental changes, spanning variations in time, weathers, and seasons. Specifically, to highlight the diversity between domains, we segment time into two prominent domains: \emph{day} and \emph{night}, with about a 12-hour time difference. The abundant daylight during the day evaluates the model's performance under regular lighting conditions, while the darker night lighting assesses its performance in low-contrast scenarios. We categorise seasons into three distinct domains—\emph{spring/summer}, \emph{autumn}, and \emph{winter}. Due to the minimal difference between spring and summer, we merge them into a single domain for convenience. Weathers are distributed across five domains, including \emph{clear}, \emph{sandstorm}, \emph{foggy}, \emph{rainy}, and \emph{snowy}. To preferably simulate realistic scenarios, \emph{snowy} only occurs in the \emph{winter} season, while the other four domains are distributed uniformly across different seasons. Notably, other weather types such as cloudy and windy are excluded due to the confined domain discrepancies.

\begin{figure*}[t]
    \centering
    \includegraphics[width=0.98\linewidth]{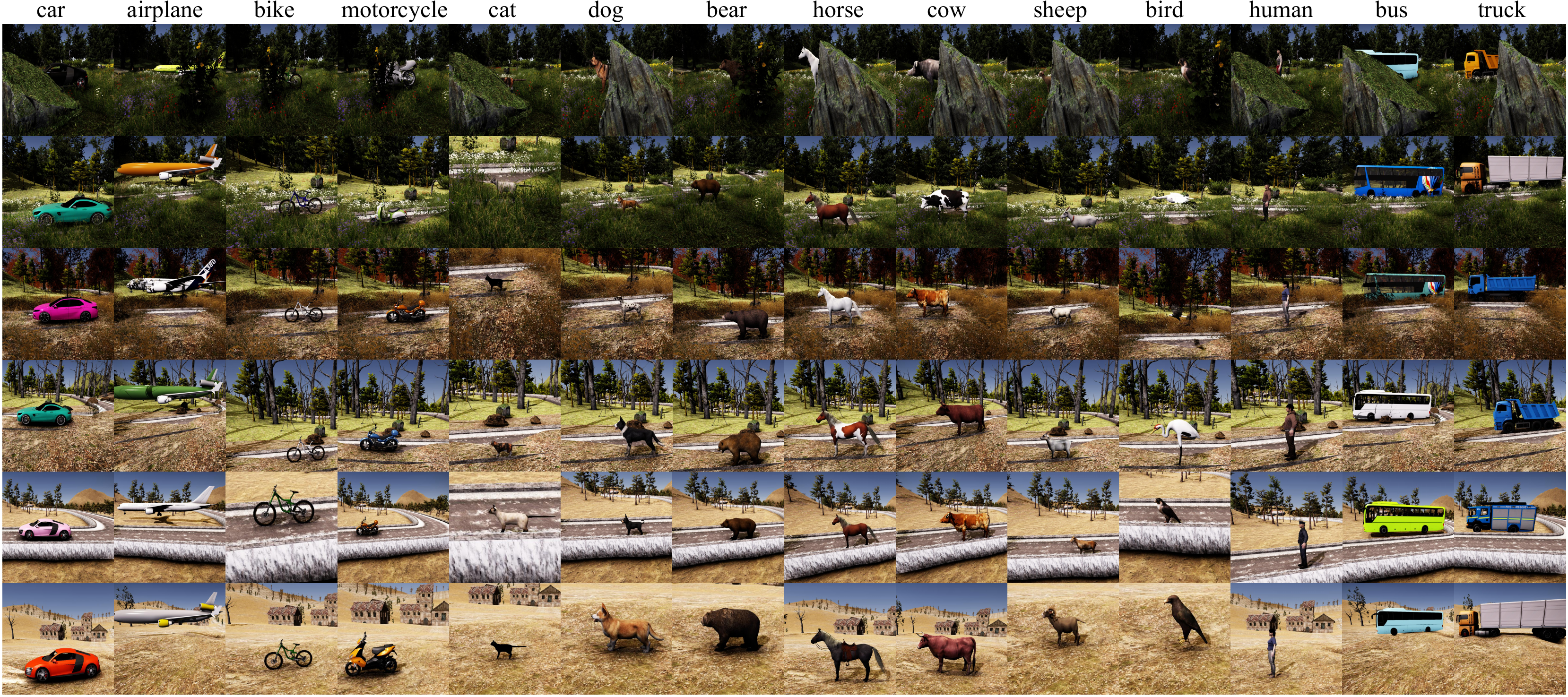}
	\caption{We showcase the DomainVerse dataset with categories arranged horizontally and domains arranged vertically. More samples can be found in the Appendix.}
	\label{fig:domainverse}
\end{figure*}

The second group contains shifts that directly impact the objects themselves, including changes in observation views and occlusions. In terms of observation views, we classify them into \emph{front} view, \emph{side (left)}  view, and \emph{top} view, motivated by the insight that three views are sufficient to fully characterize a geometric object. Notably, while the \emph{side (left)} view is a perspective frequently encountered in prior datasets, the \emph{front} and \emph{top} views are less covered but represent more challenging perspectives encountered in reality, posing difficulties for model recognition. Occlusion, a prevalent and significantly challenging domain shift in real-world scenarios, is further categorized into five domains: \emph{no occlusion}, \emph{light occlusion}, \emph{partial occlusion}, \emph{moderate occlusion}, and \emph{heavy occlusion}. Specifically, \emph{light occlusion} has an overall occlusion greater than 0 but less than 20\%, partial occlusion ranges from 20\% to 40\%, moderate occlusion from 40\% to 60\%, and heavy occlusion from 60\% to 80\%. Objects with more than 80\% occlusion are discarded due to the exorbitant recognition difficulty. To determine the mask ratio, we utilized segmentation mask tools in the game engine to calculate the ratio between generated occluded objects and unoccluded objects. 

To further increase the diversity of the proposed DomainVerse, we search several common scenes in the real world as the background of synthetic data, and manually exclude the scenes that are hard to present the above domains simultaneously. In summary, we successfully implemented a total of 5 domain shifts, corresponding to 18 coarse distinct domains, as shown in Figure~\ref{fig:domainverse}. It is worth noting that the total fine-grained domain combinations are 390 (e.g., day-autumn-clear-side-light occlusion).

\subsection{Categories}

Regarding categories, there are two groups of our DomainVerse, animals and vehicles, with 8 categories for animals and 6 categories for vehicles. The selection of these categories is guided by several considerations: 
1. The classes should predominantly appear outdoors, excluding indoor categories (e.g., sofa, stapler) and some rare categories (e.g., platypus, meerkat). 
2. The appearance of these categories in different domains should be relatively plausible. For instance, we exclude the category of a penguin as it is extremely hard to exist in a torrid \emph{summer}. 
3. The chosen classes should exhibit significant overlap with real-world datasets~\cite{everingham2010the, lin2014microsoft}, thus DomainVerse holds practical significance for real-world applications.
In summary, we feature a total of 14 categories, including \emph{car}, \emph{airplane}, \emph{bike}, \emph{motorcycle}, \emph{cat}, \emph{dog}, \emph{bear}, \emph{horse}, \emph{cow}, \emph{sheep}, \emph{bird}, \emph{human}, \emph{bus}, and \emph{truck}. 

To further enhance diversity within each category, we systematically introduce 20 different types of 3D mesh models. For vehicles, distinct mesh models represent various brands and types (e.g., Toyota, BMW, and Mercedes Benz within the \emph{car} category). For animals, different mesh models represent diverse species and variations (e.g., Labrador, Doberman, and Husky within the \emph{dog} category).

\subsection{Game Engine}
We employ Unity for data generation. To simulate weather changes, we utilize the UniStorm System (e.g., combining particle systems, mesh animation, and fog to mimic \emph{sandstorm}). For varying views, three cameras with distinct orientations are set up, simulating different perspectives with slight random angular vibrations. To represent seasons, we use multiple terrain layers to enhance scene hierarchy.

\begin{figure*}[t]
    \centering
    \includegraphics[width=0.98\linewidth]{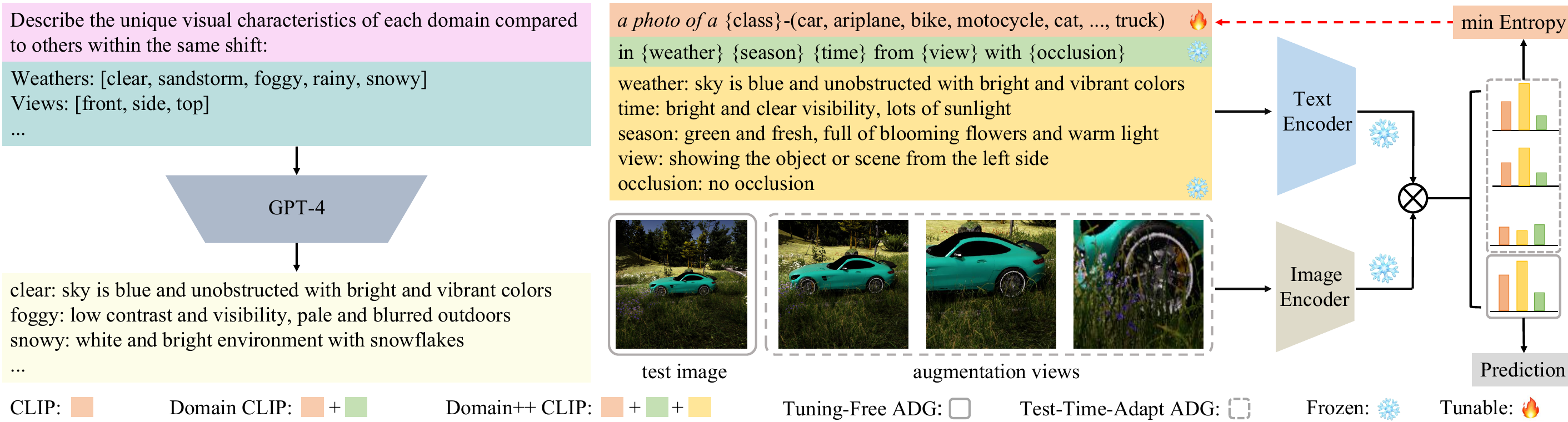}
	\caption{\textbf{Adaptive Domain Generalization}. \textbf{Left}: The generation of domain descriptions. \textbf{Right}: The pipelines of adaptative domain generalization and test-time adaptation.}
	\label{fig:domain++}
\end{figure*}
\section{Algorithm}
\label{alg}
In this section, we first introduce zero-shot CLIP as the baseline and then describe two versions of our method for tuning-free adaptive domain generalization and corresponding test-time adaptation versions.
\subsection{Zero-Shot CLIP}
CLIP employs two independent encoders: an image encoder denoted as $I()$ and a text encoder denoted as $T()$. For a downstream dataset with N categories, the conventional template ``a photo of a \{class\}" is filled with category names, serving as input to the text encoder. Subsequently, each test image undergoes encoding by the image encoder to generate the corresponding image embedding. This embedding is then compared against N text embeddings generated by the text encoder, utilizing cosine similarity as the metric. The outcome can be interpreted as the probabilities associated with various categories. Ultimately, the highest probability is selected and presented as CLIP's prediction.

\subsection{Domain-Aware CLIP}
We present two methods, named Domain CLIP and Domain++ CLIP, to incorporate prior domain knowledge into the text encoder to adapt CLIP to downstream tasks.

\noindent \textbf{Domain CLIP} The first version involves inserting the simplest form of domain information (domain name), into the standard prompt, similar to the role played by category names. Take DomainVerse as an example, where a given image may belong to domains such as \emph{light occlusion} (occlusion), \emph{side} (views), \emph{clear} (weathers), \emph{autumn} (seasons), and \emph{day} (time), the corresponding prompt input for this image is modified to ``a photo of a \{class\} in \emph{clear} \emph{autumn} \emph{day} from \emph{side} with \emph{light occlusion}." 

\noindent \textbf{Domain++ CLIP} The second version provides more detailed and distinctive features across domains. To mitigate the cost of manual annotation, we leverage a pre-trained LLM, GPT-4~\cite{achiam2023gpt}, and establish some input templates to accurately generate unique visual features of domains across domain shifts. Subsequently, we supplement these distinctive visual features into the prompt of the first version. The optimized prompt is as follows: 

``a photo of a \{class\} in \emph{clear} \emph{autumn} \emph{day} from \emph{side} with \emph{light occlusion}, \emph{sky is blue and unobstructed with bright and vibrant colors}, \emph{warm tones, crisp light, and leaves changing color}, \emph{bright and clear visibility, lots of sunlight}, \emph{showing the object or scene from the left side}, \emph{about 0 percent to 20 percent object are occluded}." 

The specific implementation details and discussion are shown in Appendix.

\subsection{Domain-Aware CLIP Algorithm for ADG}
\label{sec:ADG}
Overall, as shown in Figure~\ref{fig:domain++}, the Domain-Aware CLIP algorithm consists of two steps: (1) generating domain descriptions for each domain and (2) leveraging these domain descriptions (name) combined with the standard prompt to perform adaptive domain generalization. For the proposed tuning-free ADG, the test image is straightly put into the image encoder, and the corresponding prompts are fed to the text encoder, then the prediction is directly obtained via the cosine similarity of two embeddings. The tuning-free ADG efficiently achieve a significant improvement against the CLIP without tuning any parameters by incorporating prior domain knowledge. We also provide an alternative method named test-time-adapt ADG by fine-tuning the standard prompt (``a photo of a") based on the given test image and its different augmentation views via entropy minimization, while the domain information is frozen during tuning.

\section{Experiments}
\label{exp}
We first present the experimental setting of tuning-free adaptive domain generalization and perform extensive experiments to evaluate the baseline and current state-of-the-art (SOTA) methods on DomainVerse, with Domain CLIP and Domain++ CLIP. Then we present the performance comparison of test-time-adapt adaptive domain generalization. Next, we evaluate our method on traditional domain generalization datasets. Finally, a synthetic-to-real evaluation is introduced to prove the effectiveness of the proposed DomainVerse.

\begin{table*}[t]
\centering
\huge
\resizebox{\linewidth}{!}{
\begin{tabular}{c|c|ccc|ccc|cc}
\toprule
\multicolumn{2}{c|}{\multirow{2}{*}{Method}} & \multicolumn{3}{c|}{Zero-shot} & \multicolumn{3}{c|}{Name-only} & \multicolumn{2}{c}{Ours} \\ \cline{3-10}
\multicolumn{2}{c|}{}                                                 & \makecell[c]{CLIP\\ \cite{radford2021learning}}  & \makecell[c]{CALIP \\ \cite{guo2023calip}} & \makecell[c]{CLIP+DN \\ \cite{zhou2023distribution}} & \makecell[c]{VisDesc \\ \cite{menon2022visual}} & \makecell[c]{CuPL \\ \cite{pratt2023does}}  & \makecell[c]{CuPL+e \\ \cite{pratt2023does}}  & Domain CLIP & Domain++ CLIP\\ \midrule
\multicolumn{1}{c|}{\multirow{5}{*}{Weathers}}   & clear               & 48.30 / 56.73 & 49.79 / 57.47 & 50.14 / 56.97    
                                                                       & 46.69 / 55.13 & 47.72 / 55.56 & 48.51 / 56.60 
                                                                       & 50.64 / 58.87  & \textbf{51.00} / \textbf{59.06}   \\
\multicolumn{1}{c|}{}                           & sandstorm           & 39.90 / 49.12 & 41.05 / \textbf{50.18} & \textbf{41.96} / 49.57  
                                                                      & 37.56 / 46.91   & 38.47 / 47.45 & 39.61 / 48.90 
                                                                      & 40.68 / 48.96  & 41.27 / 50.01  \\
\multicolumn{1}{c|}{}                           & foggy               & 49.66 / 58.49 & 51.42 / 59.73 & 51.51 / 58.41
                                                                        & 48.14 / 56.60  & 49.33 / 57.94 & 50.10 / 58.88 
                                                                        & 51.94 / 60.94  & \textbf{52.38} / \textbf{61.22}    \\
\multicolumn{1}{c|}{}                           & rainy               & 48.77 / 55.92 & 49.73 / 56.40 & 50.22 / 56.39
                                                                        & 46.39 / 53.16   & 47.29 / 54.49 & 48.22 / 55.57  
                                                                        & 50.66 / 58.59  & \textbf{51.69} / \textbf{59.05}  \\
\multicolumn{1}{c|}{}                           & snowy               & 50.19 / 61.83 & 52.06 / 63.15 &  52.44 / 62.32
                                                                        & 48.89 / 58.66  & 51.21 / 60.68 & 51.34  / 62.01
                                                                        & \textbf{52.86} / \textbf{64.23}  & 52.76 / 63.19\\ \midrule
\multicolumn{1}{c|}{\multirow{3}{*}{Views}}      & front               & 48.19 / 57.61 & 49.76 / 58.34 & 49.74 / 58.42
                                                                        & 45.79 / 55.44  & 46.82 / 55.81 & 48.11 / 57.21
                                                                        & 49.88 / 58.94  & \textbf{50.96} / \textbf{59.45}  \\
\multicolumn{1}{c|}{}                           & side                & 64.37 / 72.92 & 65.51 / 73.30 & 66.38 / 72.37
                                                                        & 61.54 / 69.43   & 63.58 / 71.54 & 64.17 / 72.70 
                                                                        & 66.92 / 74.80  & \textbf{67.45} / \textbf{75.53} \\
\multicolumn{1}{c|}{}                           & top                 & 21.99 / 30.02 & 23.52 / 31.86 & \textbf{23.87} / 30.95
                                                                        & 21.79 / 29.57   & 21.72 / 29.66 & 22.47 / 30.54
                                                                        & 23.16 / \textbf{32.25}  & 23.23 / 32.04  \\ \midrule
\multicolumn{1}{c|}{\multirow{2}{*}{Time}}      & day                 & 50.65 / 59.24 & 52.26 / 60.23 & 52.25 / 59.25  
                                                                        & 49.54 / 56.62   & 50.43 / 57.49 & 51.11 / 58.98
                                                                        & \textbf{52.68} / 60.75  & 52.67 / \textbf{61.01}  \\
\multicolumn{1}{c|}{}                           & night               & 43.24 / 51.94 & 44.40 / 52.79 & 45.31 /52.52  
                                                                        & 40.53 / 50.17  & 41.86 / 51.30 & 42.87 / 52.10 
                                                                        & 44.99 / 54.08  & \textbf{46.09} / \textbf{54.58} \\ \midrule
\multicolumn{1}{c|}{\multirow{3}{*}{Seasons}}    & spring-summer       & 42.78 / 50.89 & 44.24 / 51.81 & 44.56 / 51.62
                                                                        & 41.68 / 50.14   & 42.12 / 49.67 & 42.81 / 50.69
                                                                        & 44.94 / 52.64  & \textbf{45.50} / \textbf{53.60}  \\
\multicolumn{1}{c|}{}                           & autumn              & 51.17 / 59.98 & 52.21 / 60.68 & 53.09 / 60.07
                                                                        & 48.31 / 56.72   & 49.64 / 58.74 & 50.78 / 59.94
                                                                        & 52.97 / 62.16  & \textbf{53.85} / \textbf{62.63}  \\
\multicolumn{1}{c|}{}                           & winter              & 46.67 / 55.63 & 48.29 / 56.74 & 48.48 / 55.73
                                                                        & 44.90 / 53.16   & 46.38 / 54.48 & 47.11 / 55.69
                                                                        & 48.41 / \textbf{57.21}  & \textbf{48.67} / 57.02  \\ \midrule
\multicolumn{1}{c|}{\multirow{5}{*}{Occlusion}} & no occlusion        & 56.70 / 69.53 & 58.69 / 70.28 & 58.94 / 69.69
                                                                        & 55.61 / 66.89  & 56.54 / 68.55 & 57.11 / 69.57
                                                                        & 59.38 / 71.35  & \textbf{60.48} / \textbf{72.26}  \\
\multicolumn{1}{c|}{}                           & light occlusion     & 50.41 / 58.57 & 51.24 / 59.03 & 52.64 / 58.73   
                                                                        & 47.28 / 56.03   & 48.82 / 56.95 & 49.71 / 58.08 
                                                                        & 52.56 / 60.62  & \textbf{52.97} / \textbf{61.23}  \\
\multicolumn{1}{c|}{}                           & partial occlusion   & 48.79 / 56.84 & 50.15 / 57.72 & 50.50 / 56.98   
                                                                        & 47.06 / 54.58   & 47.96 / 55.50 & 48.95 / 56.77 
                                                                        & 50.13 / 58.27  & \textbf{50.64 }/ \textbf{58.34}  \\
\multicolumn{1}{c|}{}                           & moderate occlusion  & 43.32 / 51.09 & 44.99 / 52.45 & 45.22 / 51.69  
                                                                        & 41.53 / 49.23   & 42.93 / 50.36 & 43.68 / 51.43 
                                                                        & 45.17 / 53.23  & \textbf{45.60} / \textbf{53.50}  \\
\multicolumn{1}{c|}{}                           & heavy occlusion     & 36.83 / 44.59 & 38.03 / 45.67 & 38.69 / 44.92  
                                                                        & 35.27 / 42.84 & 35.91 / 43.30 & 36.87 / 44.49 
                                                                        & 38.54 / 46.27  & \textbf{39.03} / \textbf{46.50} \\ \midrule
\multicolumn{2}{c|}{Average}                                          & 46.92 / 55.57 & 48.30 / 56.49 & 48.76 / 55.86    
                                                                            & 45.01 / 53.38   & 46.12 / 54.37 & 46.96 / 55.52 
                                                                            & 48.81 / 57.39  & \textbf{49.36} / \textbf{57.77}  \\
\bottomrule
\end{tabular}}
\caption{Performance evaluation of current works and our method on DomainVerse. Note that the performance of ViT-B/16 and ViT-L/14 are separated by $ / $.}
\label{tab:sota_eval}
\end{table*}
\subsection{Tuning-free Evaluation} 
\noindent \textbf{Dataset} For DomainVerse, the images from each fine-grained domain across different domain shifts can be seen as a distinct target domain. Due to space limitations, we only report the results from 18 domains via weighted summation.

\noindent \textbf{Experimental Settings} Inspired by~\cite{udandarao2023sus}, we evaluate current works along two major aces, zero-shot transfer and name-only transfer, neither of which requires training samples from the target task. Same as the baseline CLIP~\cite{radford2021learning}, CALIP~\cite{guo2023calip} and CLIP-DN~\cite{zhou2023distribution} adapt VLMs purely in the zero-shot setting. For CuPL~\cite{pratt2023does} and VisDesc~\cite{menon2022visual}, only the category names from the target task are needed, we use their official code to generate prompts with GPT-4~\cite{achiam2023gpt} in a preliminary phase. We subsequently report the results of the Domain CLIP and Domain++ CLIP arising from our proposed method. Notably, all the methods reported utilizing ViT-B/16 and ViT-L/14~\cite{dosovitskiy2020image} as the backbones for CLIP's image encoders, respectively.

\noindent \textbf{Main Results} As shown in Table~\ref{tab:sota_eval}, the zero-shot CLIP~\cite{radford2021learning} achieve the average accuracy of 46.92\% and 55.57\% across 5 domain shifts of DomainVerse. Specifically, \emph{sandstorm} weather performs the worst in the five domains of weather shifts due to its limited visibility, while \emph{snowy} weather achieves a better performance than \emph{clear} weather, the underlying reason is that snow eliminates the effects of other ambient noise and highlights the visual features of objects. For 3 domains of views, the \emph{side} view frequently appearing in other visual datasets makes the best accuracy of 64.37\%, and the results of \emph{front} and \emph{top} view that are rare in the pre-training stage of the VLMs drop significantly. This phenomenon further demonstrates the effectiveness of our DomainVerse for eliminating the human bias of collecting data from the \emph{side} view. The performance at different time is consistent with our expectations, the reduced visibility of \emph{night} makes recognition more difficult than the \emph{day}. Different seasons also present various results, which demonstrates that the distribution shifts of seasons in the real world are key factors for recognition. For occlusion, the accuracy decreases as the degree of occlusion increases, which is a common but easy-to-be-ignored phenomenon. 

As the results of other zero-shot methods, CALIP~\cite{guo2023calip} and CLIP-DN~\cite{zhou2023distribution} achieve an average accuracy of 48.30\% and 48.76\% on ViT-B/16. It demonstrates that post-processing embeddings based on distinct test samples is a sufficient way to transfer CLIP to downstream tasks. While for name-only methods, CuPL~\cite{pratt2023does} and VisDesc~\cite{menon2022visual}, which are designed to tackle fine-grained classification problems by introducing detailed characteristics of various objects, take a little performance decrease compared with baseline. The results demonstrate that fine-grained object attributes are less helpful for tackling realistic domain shifts. Domain CLIP outperforms the baseline by 1.89\% and 1.82\% on ViT-B/16 and ViT-L/14, respectively. Besides, Domain++ CLIP achieves a further improvement with SOTA accuracies of 49.36\% and 57.77\%, which represents the 
the essential of prior domain knowledge for ADG.

\subsection{Test-time-adapt Evaluation}
\begin{table*}[t]
\huge
\centering
\resizebox{\linewidth}{!}{
\begin{tabular}{c|c|ccccc|ccc|cc|ccc|ccccc}
\toprule
 \multirow{2}{*}{Method} & \multirow{2}{*}{Average} & \multicolumn{5}{c|}{Weathers} & \multicolumn{3}{c|}{Views} & \multicolumn{2}{c|}{Time} & \multicolumn{3}{c|}{Seasons} &\multicolumn{5}{c}{Occlusion} \\ \cline{3-20}
 & & clear & sand & foggy & rainy & snowy & front & side & top & day & night & spring & autumn & winter & no & light & partial &moderate & heavy \\ \midrule
 Baseline & 46.92 & 48.30 & 39.90 & 49.66 & 48.77 & 50.19 & 48.19 & 64.37 & 21.99 & 50.65 & 43.24 & 42.78 & 51.17 & 46.67 & 56.70 & 50.41 & 48.79 & 43.32 & 36.83 \\
TPT~\cite{shu2022test} & 49.94 & 51.48 & 43.25 & 52.70 & 51.12 & 53.73 & 52.01 & 67.25 & 24.44 & 53.53 & 46.41 & 45.46 & 54.10 & 50.07 & 59.71 & 53.20 & 51.96 & 46.48 & 39.74 \\
Domain CLIP-T & \textbf{53.11} & \textbf{55.68} & \textbf{45.13} & \textbf{56.11} & \textbf{54.40} & \textbf{56.60} & \textbf{55.93} & \textbf{70.75}& \textbf{26.42} & \textbf{56.50} & \textbf{49.77} & \textbf{49.49} & \textbf{57.11} & \textbf{52.65} & \textbf{65.17} & \textbf{56.29} & \textbf{54.22} & \textbf{49.46} & \textbf{42.62} \\
Domain++ CLIP-T & 52.59 & 54.82 & 45.11 & 55.26 & 54.25 & 55.42 & 55.21 & 70.30 & 25.99 & 55.71 & 49.51 & 49.16 & 56.49  & 52.06  & 64.32 & \textbf{56.29} & 53.63 & 48.77 & 42.04  \\ \midrule \midrule
Baseline & 55.57 & 56.73 & 49.12 & 58.49 & 55.92 & 61.83 & 57.61 & 72.92 & 30.02 & 59.24 & 51.94 & 50.89 & 59.98 & 55.63 & 69.53 & 58.57 & 56.84 & 51.09 & 44.59 \\
TPT~\cite{shu2022test} & 59.80 & 58.49 & 54.13 & \textbf{66.23} & 56.97 & \textbf{68.52} & 61.79 & 76.81 & 33.75 & 63.46 & 56.24 & 56.14 & 63.23 & 60.17 & 72.23 & 61.45 & \textbf{62.61} & 56.52 & 48.42 \\
Domain CLIP-T & 61.26 & 61.82 & \textbf{54.43} & 66.08 & 59.72 & 68.28 & 63.29 & 77.96 & \textbf{37.01} & 64.06 & \textbf{58.58} & 57.23 & 64.66& \textbf{61.81}& 75.12 & 64.33 & 61.88 & \textbf{57.44} & 50.07 \\
Domain++ CLIP-T & \textbf{61.62} & \textbf{63.17} & 54.20 & 65.01 & \textbf{62.34} & 67.09 & \textbf{64.48} & \textbf{78.70} & 35.66 & \textbf{64.85} & 58.44 & \textbf{58.00} & \textbf{65.73}  & 61.07  & \textbf{76.42} & \textbf{65.05} & 62.09 & \textbf{57.44} & \textbf{50.16}  \\
\bottomrule
\end{tabular}}
\caption{Test-time-adapt ADG Evaluation on DomainVerse. Note that \emph{Domain-aware CLIP-T} is used to distinguish it from Tuning-free ADG as described in Section~\ref{sec:ADG}. The upper part of the table is the results on ViT-B/16, and the lower part is for ViT-L/14. Due to space limitations, some domain names are replaced by abbreviations (sand: sandstorm, no: no occlusion, light: light occlusion, partial: partial occlusion, moderate: moderate occlusion, heavy: heavy occlusion).}
\label{tab:tpt1}
\end{table*}
\noindent \textbf{Experimental Settings} We follow the standard test-time adaptation strategy~\cite{shu2022test} that tunes the parameters (prompt for CLIP-based models) on the fly for each test sample. Specifically, a batch of 64 images, including an original test image and its 63 augmentations, is put into the pre-trained model to optimize the prompt via entropy minimization. For fair evaluation, the initial prefix of prompts, ``a photo of a", is optimized during test time. While the domain information is fixed during inference as the prior knowledge for our method.

\noindent \textbf{Main Results} The Table~\ref{tab:tpt1} presents the performance comparison of TPT~\cite{shu2022test} and our methods. It can be seen that test-time adaptation achieves a remarked improvement compared with the baseline. For our methods, Domain CLIP-T achieves a better performance on most of the domains and significant improvement for average accuracy of 3.17\% and 2.65\% compared with TPT, respectively. Although Domain++ CLIP-T fails to make a further improvement compared with Domain CLIP-T under ViT-B/16, the performance is much better than TPT and its tuning-free version. The underlying reason may lie in that fine-tuning the long-series prompts is a challenge for current VLMs. For ViT-L/14, Domain++ CLIP-T is slightly better than Domain CLIP-T, which may be attributed to the stronger ability to deal with long-series text prompts of a larger backbone. In general, tuning domain-independent prompts with fixed domain-aware prompts is more effective for test-time adaptation in complex realistic scenarios.

\subsection{Traditional DG Evaluation}
\begin{table}[t]
\resizebox{\linewidth}{!}{
\begin{tabular}{c|cccc|c}
\toprule
\multicolumn{6}{c}{\emph{\textbf{PACS}}} \\
\midrule
Methods & Art  & Cartoon & Photo & Sketch & Avg.           \\ \midrule
Baseline~\cite{radford2021learning}  & 97.22  & 99.10 & \textbf{99.94} &88.24 &94.58 \\
Domain CLIP (ours) & 97.61 & \textbf{99.40} & \textbf{99.94} & 90.63 & 95.68 \\
Domain++ CLIP (ours) & \textbf{97.80} & \textbf{99.40} & 99.88 & \textbf{91.80} & \textbf{96.17} \\
\midrule 
\multicolumn{6}{c}{\emph{\textbf{Office-Home}}} \\
\midrule
Methods & Art  & Clipart & Product & Real-world & Avg.           \\ \midrule
Baseline~\cite{radford2021learning}  & 79.36  & 66.92 & 87.95 &88.50 &80.88 \\
Domain CLIP (ours) & \textbf{80.22} & 69.81 & 89.75 & \textbf{89.37} & 82.58 \\
Domain++ CLIP (ours) & 79.98 & \textbf{71.43} & \textbf{89.79} & 88.87 & \textbf{82.87}
\\ \bottomrule
\end{tabular}}
\caption{Traditional domain generalization evaluation. Note that zero-shot CLIP~\cite{radford2021learning} utilizing the standard prompt ``a photo of a [class]" is the baseline. For the Domain CLIP, ``a \{\} of a [class]" filled with the target domain name except for the photo domain is leveraged as the prompt. The Domain++ CLIP adds a detailed domain description compared with the Domain CLIP.}
\label{tab:pacs}
\end{table}
\noindent \textbf{Dataset} PACS~\cite{li2017deeper} is composed of four distinct domains: \emph{art painting}, \emph{cartoon}, \emph{photo}, and \emph{sketch}. Each domain exhibits different image styles, and there are seven classes in total. The dataset comprises a total of 9,991 images, each with a resolution of $227\times 227$. Office-Home~\cite{venkateswara2017deep} is a more challenging dataset, containing four domains: \emph{art}, \emph{clipart}, \emph{product}, and \emph{real-world}. Each domain contains 65 categories commonly found in office and home settings. The two datasets are the most popular benchmarks for DG.

\noindent \textbf{Experimental Settings} Different from the leave-one-domain-out evaluation strategy of traditional domain generalization, selecting three domains as the source for training, and the remaining domain as the target for evaluation~\cite{zhou2023mixstyle}. We leverage the CLIP~\cite{radford2021learning} model as the source model and adapt it to four domains from PACS~\cite{li2017deeper} and Office-Home~\cite{venkateswara2017deep}, performing a zero-shot evaluation. The zero-shot CLIP is the baseline and compares against Domain CLIP and Domain++ CLIP. We use the ResNet-50~\cite{he2016deep} as the backbone of CLIP's image encoder.

\noindent \textbf{Main Results} As shown in Table~\ref{tab:pacs}, the vanilla CLIP~\cite{radford2021learning} serves a strong baseline, achieving an average accuracy of 94.58\% on PACS~\cite{li2017deeper}. Our method showcases even more impressive results, particularly excelling in 3 domains: \emph{art-painting}, \emph{cartoon}, and \emph{sketch}. With superior performance in these domains, our approach achieves SOTA average accuracies of 95.68\% and 96.17\%, respectively. For Office-Home~\cite{venkateswara2017deep}, our method outperforms the baseline with average accuracies of 1.70\% and 1.99\%. This significant improvement demonstrates that our method performs well both on real datasets and classical datasets with various domains.

\subsection{Synthetic to Real Evaluation}
\noindent \textbf{Dataset} We gather real-world images from the VOC2007 and MS-COCO datasets \cite{everingham2010the, lin2014microsoft}, focusing on categories that significantly overlap with DomainVerse. As these datasets primarily cater to multi-label classification, we filter out images containing only one of the 14 classes present in DomainVerse. Subsequently, 68,671 images meeting these criteria are selected as the real-world dataset for this section, called DWild.

\begin{table}[t]
\centering
\resizebox{0.9\linewidth}{!}{
\begin{tabular}{c|cc|c}
\toprule
Models & Baseline  & TPT & DomainVerse TPT (ours)    \\ \midrule
ViT-B/16  & 72.20 / 93.63  & 74.51 / 92.94 & \textbf{80.51} / \textbf{96.21} \\
\bottomrule
\end{tabular}}
\caption{Synthetic to Real Evaluation. Both top1 and top5 accuracy on ViT-B/16 are reported.}
\label{tab:syn2real}
\end{table}

\noindent \textbf{Experimental Settings} To better evaluate the practicality of the introduced DomainVerse, we experiment to transfer the model trained on DomainVerse to a real dataset, DWild. We first follow the popular prompt fine-tuning method~\cite{zhou2022learning} to optimize prompts on DomainVerse. For fair evaluation, the standard initial prompt ``a photo of a \{class\}" is optimized during training on DomainVerse, ignoring its diverse domain information. Finally, the trained model is utilized to perform test-time adaptation on DWild.

\noindent \textbf{Main Results} As shown in Table~\ref{tab:syn2real}, compared with the baseline that leverages CLIP~\cite{radford2021learning} to perform zero-shot generalization on DWild, TPT~\cite{shu2022test} achieves a better result with 74.51\% on top1 accuracy, while worse performance on top5 accuracy. After training on the proposed DomainVerse and optimizing the initial prompts, the TPT method makes a significant improvement of 6.00\% on top1 accuracy and 3.27\% on top5 accuracy, respectively. It demonstrates that the prompts optimized on DomainVerse offer a better initialization than the standard initial strategy. In other words, the proposed DomainVerse can be utilized as a pre-trained dataset to tackle real-world distribution shifts.

\section{Conclusion}
\label{sec:conclusion}
We construct a novel large-scale dataset named DomainVerse to facilitate the VLMs to achieve adaptive domain generalization on realistic scenarios. Specifically, DomainVerse consists of 5 common domain shifts in the real world and corresponding 18 coarse domains with a balanced collection manner. Moreover, Domain CLIP and Domain++ CLIP are proposed to achieve tuning-free adaptive domain generalization by leveraging the prior domain knowledge. Extensive experiments demonstrate the significance of the DomainVerse and the effectiveness of the our methods.

\nocite{langley00}

\bibliography{0-main}

\begin{thebibliography}{41}
\providecommand{\natexlab}[1]{#1}
\providecommand{\url}[1]{\texttt{#1}}
\expandafter\ifx\csname urlstyle\endcsname\relax
  \providecommand{\doi}[1]{doi: #1}\else
  \providecommand{\doi}{doi: \begingroup \urlstyle{rm}\Url}\fi

\bibitem[Achiam et~al.(2023)Achiam, Adler, Agarwal, Ahmad, Akkaya, Aleman, Almeida, Altenschmidt, Altman, Anadkat, et~al.]{achiam2023gpt}
Achiam, J., Adler, S., Agarwal, S., Ahmad, L., Akkaya, I., Aleman, F.~L., Almeida, D., Altenschmidt, J., Altman, S., Anadkat, S., et~al.
\newblock Gpt-4 technical report.
\newblock \emph{arXiv preprint arXiv:2303.08774}, 2023.

\bibitem[Chen et~al.(2023)Chen, Zhang, Han, Chen, Shi, Xu, and Xu]{chen2023vlp}
Chen, F.-L., Zhang, D.-Z., Han, M.-L., Chen, X.-Y., Shi, J., Xu, S., and Xu, B.
\newblock Vlp: A survey on vision-language pre-training.
\newblock \emph{Machine Intelligence Research}, 20\penalty0 (1):\penalty0 38--56, 2023.

\bibitem[Dosovitskiy et~al.(2020)Dosovitskiy, Beyer, Kolesnikov, Weissenborn, Zhai, Unterthiner, Dehghani, Minderer, Heigold, Gelly, et~al.]{dosovitskiy2020image}
Dosovitskiy, A., Beyer, L., Kolesnikov, A., Weissenborn, D., Zhai, X., Unterthiner, T., Dehghani, M., Minderer, M., Heigold, G., Gelly, S., et~al.
\newblock An image is worth 16x16 words: Transformers for image recognition at scale.
\newblock \emph{arXiv preprint arXiv:2010.11929}, 2020.

\bibitem[Du et~al.(2022)Du, Liu, Li, and Zhao]{du2022survey}
Du, Y., Liu, Z., Li, J., and Zhao, W.~X.
\newblock A survey of vision-language pre-trained models.
\newblock \emph{arXiv preprint arXiv:2202.10936}, 2022.

\bibitem[{Everingham} et~al.(2010){Everingham}, {Gool}, {Williams}, {Winn}, and {Zisserman}]{everingham2010the}
{Everingham}, M., {Gool}, L., {Williams}, C.~K., {Winn}, J., and {Zisserman}, A.
\newblock The pascal visual object classes (voc) challenge.
\newblock \emph{International Journal of Computer Vision}, 88\penalty0 (2):\penalty0 303--338, 2010.

\bibitem[Ganin et~al.(2016)Ganin, Ustinova, Ajakan, Germain, Larochelle, Laviolette, Marchand, and Lempitsky]{ganin2016domain}
Ganin, Y., Ustinova, E., Ajakan, H., Germain, P., Larochelle, H., Laviolette, F., Marchand, M., and Lempitsky, V.
\newblock Domain-adversarial training of neural networks.
\newblock \emph{The journal of machine learning research}, 17\penalty0 (1):\penalty0 2096--2030, 2016.

\bibitem[Gong et~al.(2012)Gong, Shi, Sha, and Grauman]{gong2012geodesic}
Gong, B., Shi, Y., Sha, F., and Grauman, K.
\newblock Geodesic flow kernel for unsupervised domain adaptation.
\newblock In \emph{2012 IEEE conference on computer vision and pattern recognition}, pp.\  2066--2073. IEEE, 2012.

\bibitem[Guo et~al.(2023)Guo, Zhang, Qiu, Ma, Miao, He, and Cui]{guo2023calip}
Guo, Z., Zhang, R., Qiu, L., Ma, X., Miao, X., He, X., and Cui, B.
\newblock Calip: Zero-shot enhancement of clip with parameter-free attention.
\newblock In \emph{Proceedings of the AAAI Conference on Artificial Intelligence}, volume~37, pp.\  746--754, 2023.

\bibitem[Haas(2014)]{haas2014history}
Haas, J.~K.
\newblock A history of the unity game engine.
\newblock \emph{Diss. Worcester Polytechnic Institute}, 483\penalty0 (2014):\penalty0 484, 2014.

\bibitem[He et~al.(2016)He, Zhang, Ren, and Sun]{he2016deep}
He, K., Zhang, X., Ren, S., and Sun, J.
\newblock Deep residual learning for image recognition.
\newblock In \emph{Proceedings of the IEEE conference on computer vision and pattern recognition}, pp.\  770--778, 2016.

\bibitem[Hou et~al.(2023)Hou, Zhang, Liu, Yuan, Zhong, Zhang, Shi, Fan, and He]{hou2023learning}
Hou, F., Zhang, Y., Liu, Y., Yuan, J., Zhong, C., Zhang, Y., Shi, Z., Fan, J., and He, Z.
\newblock Learning how to learn domain-invariant parameters for domain generalization.
\newblock In \emph{ICASSP 2023-2023 IEEE International Conference on Acoustics, Speech and Signal Processing (ICASSP)}, pp.\  1--5. IEEE, 2023.

\bibitem[Hull(1994)]{hull1994database}
Hull, J.~J.
\newblock A database for handwritten text recognition research.
\newblock \emph{IEEE Transactions on pattern analysis and machine intelligence}, 16\penalty0 (5):\penalty0 550--554, 1994.

\bibitem[Hussain et~al.(2020)Hussain, Shakeel, Hussain, Uddin, and Ghouri]{hussain2020unity}
Hussain, A., Shakeel, H., Hussain, F., Uddin, N., and Ghouri, T.~L.
\newblock Unity game development engine: A technical survey.
\newblock \emph{Univ. Sindh J. Inf. Commun. Technol}, 4:\penalty0 73--81, 2020.

\bibitem[Jia et~al.(2021)Jia, Yang, Xia, Chen, Parekh, Pham, Le, Sung, Li, and Duerig]{jia2021scaling}
Jia, C., Yang, Y., Xia, Y., Chen, Y.-T., Parekh, Z., Pham, H., Le, Q., Sung, Y.-H., Li, Z., and Duerig, T.
\newblock Scaling up visual and vision-language representation learning with noisy text supervision.
\newblock In \emph{International conference on machine learning}, pp.\  4904--4916. PMLR, 2021.

\bibitem[Kirillov et~al.(2023)Kirillov, Mintun, Ravi, Mao, Rolland, Gustafson, Xiao, Whitehead, Berg, Lo, et~al.]{kirillov2023segment}
Kirillov, A., Mintun, E., Ravi, N., Mao, H., Rolland, C., Gustafson, L., Xiao, T., Whitehead, S., Berg, A.~C., Lo, W.-Y., et~al.
\newblock Segment anything.
\newblock \emph{arXiv preprint arXiv:2304.02643}, 2023.

\bibitem[LeCun et~al.(1998)LeCun, Bottou, Bengio, and Haffner]{lecun1998gradient}
LeCun, Y., Bottou, L., Bengio, Y., and Haffner, P.
\newblock Gradient-based learning applied to document recognition.
\newblock \emph{Proceedings of the IEEE}, 86\penalty0 (11):\penalty0 2278--2324, 1998.

\bibitem[Li et~al.(2017)Li, Yang, Song, and Hospedales]{li2017deeper}
Li, D., Yang, Y., Song, Y.-Z., and Hospedales, T.~M.
\newblock Deeper, broader and artier domain generalization.
\newblock In \emph{Proceedings of the IEEE international conference on computer vision}, pp.\  5542--5550, 2017.

\bibitem[{Lin} et~al.(2014){Lin}, {Maire}, {Belongie}, {Hays}, {Perona}, {Ramanan}, {Dollár}, and {Zitnick}]{lin2014microsoft}
{Lin}, T.-Y., {Maire}, M., {Belongie}, S.~J., {Hays}, J., {Perona}, P., {Ramanan}, D., {Dollár}, P., and {Zitnick}, C.~L.
\newblock Microsoft coco: Common objects in context.
\newblock In \emph{European Conference on Computer Vision}, pp.\  740--755, 2014.

\bibitem[Menon \& Vondrick(2022)Menon and Vondrick]{menon2022visual}
Menon, S. and Vondrick, C.
\newblock Visual classification via description from large language models.
\newblock \emph{arXiv preprint arXiv:2210.07183}, 2022.

\bibitem[Muralidhar et~al.(2018)Muralidhar, Islam, Marwah, Karpatne, and Ramakrishnan]{muralidhar2018incorporating}
Muralidhar, N., Islam, M.~R., Marwah, M., Karpatne, A., and Ramakrishnan, N.
\newblock Incorporating prior domain knowledge into deep neural networks.
\newblock In \emph{2018 IEEE international conference on big data (big data)}, pp.\  36--45. IEEE, 2018.

\bibitem[Netzer et~al.(2011)Netzer, Wang, Coates, Bissacco, Wu, and Ng]{netzer2011reading}
Netzer, Y., Wang, T., Coates, A., Bissacco, A., Wu, B., and Ng, A.~Y.
\newblock Reading digits in natural images with unsupervised feature learning.
\newblock 2011.

\bibitem[Peng et~al.(2019)Peng, Bai, Xia, Huang, Saenko, and Wang]{peng2019moment}
Peng, X., Bai, Q., Xia, X., Huang, Z., Saenko, K., and Wang, B.
\newblock Moment matching for multi-source domain adaptation.
\newblock In \emph{Proceedings of the IEEE/CVF international conference on computer vision}, pp.\  1406--1415, 2019.

\bibitem[Pratt et~al.(2023)Pratt, Covert, Liu, and Farhadi]{pratt2023does}
Pratt, S., Covert, I., Liu, R., and Farhadi, A.
\newblock What does a platypus look like? generating customized prompts for zero-shot image classification.
\newblock In \emph{Proceedings of the IEEE/CVF International Conference on Computer Vision}, pp.\  15691--15701, 2023.

\bibitem[Radford et~al.(2021)Radford, Kim, Hallacy, Ramesh, Goh, Agarwal, Sastry, Askell, Mishkin, Clark, et~al.]{radford2021learning}
Radford, A., Kim, J.~W., Hallacy, C., Ramesh, A., Goh, G., Agarwal, S., Sastry, G., Askell, A., Mishkin, P., Clark, J., et~al.
\newblock Learning transferable visual models from natural language supervision.
\newblock In \emph{International conference on machine learning}, pp.\  8748--8763. PMLR, 2021.

\bibitem[Saenko et~al.(2010)Saenko, Kulis, Fritz, and Darrell]{saenko2010adapting}
Saenko, K., Kulis, B., Fritz, M., and Darrell, T.
\newblock Adapting visual category models to new domains.
\newblock In \emph{European conference on computer vision}, pp.\  213--226. Springer, 2010.

\bibitem[Shu et~al.(2022)Shu, Nie, Huang, Yu, Goldstein, Anandkumar, and Xiao]{shu2022test}
Shu, M., Nie, W., Huang, D.-A., Yu, Z., Goldstein, T., Anandkumar, A., and Xiao, C.
\newblock Test-time prompt tuning for zero-shot generalization in vision-language models.
\newblock \emph{Advances in Neural Information Processing Systems}, 35:\penalty0 14274--14289, 2022.

\bibitem[Singh et~al.(2022)Singh, Hu, Goswami, Couairon, Galuba, Rohrbach, and Kiela]{singh2022flava}
Singh, A., Hu, R., Goswami, V., Couairon, G., Galuba, W., Rohrbach, M., and Kiela, D.
\newblock Flava: A foundational language and vision alignment model.
\newblock In \emph{Proceedings of the IEEE/CVF Conference on Computer Vision and Pattern Recognition}, pp.\  15638--15650, 2022.

\bibitem[Torralba \& Efros(2011)Torralba and Efros]{torralba2011unbiased}
Torralba, A. and Efros, A.~A.
\newblock Unbiased look at dataset bias.
\newblock In \emph{CVPR 2011}, pp.\  1521--1528. IEEE, 2011.

\bibitem[Udandarao et~al.(2023)Udandarao, Gupta, and Albanie]{udandarao2023sus}
Udandarao, V., Gupta, A., and Albanie, S.
\newblock Sus-x: Training-free name-only transfer of vision-language models.
\newblock In \emph{Proceedings of the IEEE/CVF International Conference on Computer Vision}, pp.\  2725--2736, 2023.

\bibitem[Venkateswara et~al.(2017)Venkateswara, Eusebio, Chakraborty, and Panchanathan]{venkateswara2017deep}
Venkateswara, H., Eusebio, J., Chakraborty, S., and Panchanathan, S.
\newblock Deep hashing network for unsupervised domain adaptation.
\newblock In \emph{Proceedings of the IEEE conference on computer vision and pattern recognition}, pp.\  5018--5027, 2017.

\bibitem[Xu et~al.(2021)Xu, Zhang, Zhang, Wang, et~al.]{QinweiXu2021AFF}
Xu, Q., Zhang, R., Zhang, Y., Wang, Y., et~al.
\newblock A fourier-based framework for domain generalization.
\newblock In \emph{CVPR}, 2021.

\bibitem[Yuan et~al.(2022)Yuan, Hou, Du, Shi, Geng, Fan, and Rui]{yuan2022self}
Yuan, J., Hou, F., Du, Y., Shi, Z., Geng, X., Fan, J., and Rui, Y.
\newblock Self-supervised graph neural network for multi-source domain adaptation.
\newblock In \emph{Proceedings of the 30th ACM International Conference on Multimedia}, pp.\  3907--3916, 2022.

\bibitem[Zhang et~al.(2023)Zhang, He, Xu, Yu, Shen, and Cui]{zhang2023nico++}
Zhang, X., He, Y., Xu, R., Yu, H., Shen, Z., and Cui, P.
\newblock Nico++: Towards better benchmarking for domain generalization.
\newblock In \emph{Proceedings of the IEEE/CVF Conference on Computer Vision and Pattern Recognition}, pp.\  16036--16047, 2023.

\bibitem[Zhao et~al.(2022)Zhao, Yu, Ma, Yu, Mei, Wang, He, Yuille, and Kortylewski]{zhao2022ood}
Zhao, B., Yu, S., Ma, W., Yu, M., Mei, S., Wang, A., He, J., Yuille, A., and Kortylewski, A.
\newblock Ood-cv: a benchmark for robustness to out-of-distribution shifts of individual nuisances in natural images.
\newblock In \emph{European Conference on Computer Vision}, pp.\  163--180. Springer, 2022.

\bibitem[Zhong et~al.(2022)Zhong, Yang, Zhang, Li, Codella, Li, Zhou, Dai, Yuan, Li, et~al.]{zhong2022regionclip}
Zhong, Y., Yang, J., Zhang, P., Li, C., Codella, N., Li, L.~H., Zhou, L., Dai, X., Yuan, L., Li, Y., et~al.
\newblock Regionclip: Region-based language-image pretraining.
\newblock In \emph{Proceedings of the IEEE/CVF Conference on Computer Vision and Pattern Recognition}, pp.\  16793--16803, 2022.

\bibitem[Zhou et~al.(2021{\natexlab{a}})Zhou, Liu, Qiao, Xiang, et~al.]{zhou2021domainsuvery}
Zhou, K., Liu, Z., Qiao, Y., Xiang, T., et~al.
\newblock Domain generalization: A survey.
\newblock \emph{arXiv preprint arXiv:2103.02503}, 2021{\natexlab{a}}.

\bibitem[Zhou et~al.(2021{\natexlab{b}})Zhou, Yang, Qiao, and Xiang]{zhou2021domain}
Zhou, K., Yang, Y., Qiao, Y., and Xiang, T.
\newblock Domain adaptive ensemble learning.
\newblock \emph{IEEE Transactions on Image Processing}, 30:\penalty0 8008--8018, 2021{\natexlab{b}}.

\bibitem[Zhou et~al.(2022{\natexlab{a}})Zhou, Yang, Loy, and Liu]{zhou2022conditional}
Zhou, K., Yang, J., Loy, C.~C., and Liu, Z.
\newblock Conditional prompt learning for vision-language models.
\newblock In \emph{Proceedings of the IEEE/CVF Conference on Computer Vision and Pattern Recognition}, pp.\  16816--16825, 2022{\natexlab{a}}.

\bibitem[Zhou et~al.(2022{\natexlab{b}})Zhou, Yang, Loy, and Liu]{zhou2022learning}
Zhou, K., Yang, J., Loy, C.~C., and Liu, Z.
\newblock Learning to prompt for vision-language models.
\newblock \emph{International Journal of Computer Vision}, 130\penalty0 (9):\penalty0 2337--2348, 2022{\natexlab{b}}.

\bibitem[Zhou et~al.(2023{\natexlab{a}})Zhou, Yang, Qiao, and Xiang]{zhou2023mixstyle}
Zhou, K., Yang, Y., Qiao, Y., and Xiang, T.
\newblock Mixstyle neural networks for domain generalization and adaptation.
\newblock \emph{International Journal of Computer Vision}, pp.\  1--15, 2023{\natexlab{a}}.

\bibitem[Zhou et~al.(2023{\natexlab{b}})Zhou, Ren, Li, Zabih, and Lim]{zhou2023distribution}
Zhou, Y., Ren, J., Li, F., Zabih, R., and Lim, S.-N.
\newblock Distribution normalization: An" effortless" test-time augmentation for contrastively learned visual-language models.
\newblock \emph{arXiv preprint arXiv:2302.11084}, 2023{\natexlab{b}}.

\end{thebibliography}
\bibliographystyle{icml2023}

\end{document}